\documentclass[letterpaper, 10 pt, journal, twoside]{IEEEtran}
\usepackage{graphicx}
\usepackage{amsmath}
\usepackage{amssymb}
\usepackage{booktabs}
\usepackage[table]{xcolor}
\usepackage{multirow}
\usepackage{array, makecell}
\usepackage{lettrine}
\usepackage{ulem}
\usepackage{caption}
\usepackage{subcaption}
\usepackage{changes}
\definechangesauthor[name={Sandika}, color=orange]{sb}
\usepackage[top=57pt,bottom=43pt,left=48pt,right=57pt]{geometry}

\ifCLASSINFOpdf
\else
\fi

\makeatletter
\let\NAT@parse\undefined
\makeatother
\usepackage{hyperref}
\hypersetup{citecolor=green,
            linkcolor=red}

\def\eg{\emph{e.g}., } 
\def\ie{\emph{i.e}., } 
 
\def\etc{\emph{etc}.} 

\def\wrt{\emph{w.r.t.}~}

\def\etal{\emph{et al}.~}

\begin{document}
%
\title{\LARGE \bf Physically Plausible 3D Human-Scene Reconstruction from Monocular RGB Image using an Adversarial Learning Approach}
%
%
%

\author{Sandika Biswas$^{1,3}$, Kejie Li$^2$, Biplab Banerjee$^3$, Subhasis Chaudhuri$^3$, Hamid Rezatofighi$^1$%
\thanks{Manuscript received: March, 11, 2023; Revised: June, 5, 2023; Accepted:July, 5, 2023.}
\thanks{$^{1,3}$Sandika Biswas is with the Indian Institute of Technology, Bombay, and Monash University. Email:
       {\tt\small sandika.biswas@monash.edu}}%
\thanks{$^2$Kejie Li is with the University of Oxford. Email:
       {\tt\small kejie.li@outlook.com}}%
\thanks{$^3$Biplab Banerjee and Subhasis Chaudhuri are with the Indian Institute of Technology, Bombay. Email:
       {\tt\small bbanerjee@iitb.ac.in}, {\tt\small sc@iitb.ac.in}} 
\thanks{$^1$Hamid Rezatofighi is with the Faculty of Information Technology, Monash University. Email:
       {\tt\small hamid.rezatofighi@monash.edu}}
\thanks{Digital Object Identifier (DOI): see top of this page.}
}
%
%

\markboth{IEEE Robotics and Automation Letters. Preprint Version. Accepted JULY, 2023}
{Biswas \MakeLowercase{\textit{et al.}}: 3D Human-Scene Reconstruction using Adversarial Learning} 

%



\maketitle

\begin{abstract}
Holistic 3D human-scene reconstruction is a crucial and emerging research area in robot perception. A key challenge in holistic 3D human-scene reconstruction is to generate a physically plausible 3D scene from a single monocular RGB image. 
The existing research mainly proposes optimization-based approaches for reconstructing the scene from a sequence of RGB frames with explicitly defined physical laws and constraints between different scene elements (humans and objects). However, it is hard to explicitly define and model every physical law in every scenario. This paper proposes using an implicit feature representation of the scene elements to distinguish a physically plausible alignment of humans and objects from an implausible one. We propose using a graph-based holistic representation with an encoded physical representation of the scene to analyze the human-object and object-object interactions within the scene. Using this graphical representation, we adversarially train our model to learn the feasible alignments of the scene elements from the training data itself without explicitly defining the laws and constraints between them. Unlike the existing inference-time optimization-based approaches, we use this adversarially trained model to
produce a per-frame 3D reconstruction of the scene that abides by the physical laws and constraints. Our learning-based method achieves comparable 3D reconstruction quality to existing optimization-based holistic human-scene reconstruction methods and does not need inference time optimization. This makes it better suited when compared to existing methods, for potential use in robotic applications, such as robot navigation, \etc 
\end{abstract}


%
\IEEEpeerreviewmaketitle

\section{Introduction}
\IEEEPARstart{H}{uman}-scene reconstruction in 3D is crucial for successful robot navigation and interactions in the 3D world. Robots need to perceive and understand the surrounding environment to navigate and interact effectively. From the 3D reconstruction of the scene, robots are able to gain a more comprehensive understanding of their surroundings, including the shapes, positions, and spatial relationships of objects. To make informed decisions and plan its actions robot should have an accurate knowledge of the scene structure, hence physically plausible reconstruction is necessary for robot navigation.
Significant progress has been made in the independent reconstruction of the humans \cite{zhang2021holistic,choutas2020monocular,kocabas2020vibe}, \cite{kanazawa2018end,rong2021frankmocap,kocabas2021pare}, and the objects \cite{dahnert2021panoptic,gkioxari2019mesh,Nie_2020_CVPR},
\cite{song2017semantic,nie2022learning} present in a scene. But only a few recent works \cite{weng2021holistic,yi2022human} focus on holistic 3D human-scene reconstruction. 

The human and object elements of a scene constrain each other's positions, orientations, and alignments to avoid a collision or to ensure contact while interacting with each other. Hence, for the holistic scene reconstruction system, it is necessary to learn these constraints based on the human-object (\eg a person sitting on a  chair) and object-object (\eg a chair in front of a table) relations or interactions for physically plausible reconstruction. The scene reconstruction system should also ensure the physical laws in the produced reconstruction \eg humans/objects can not float over the floor or humans cannot walk through the wall \etc

\begin{figure}[t]
\centering
 \includegraphics[width=0.8\columnwidth]{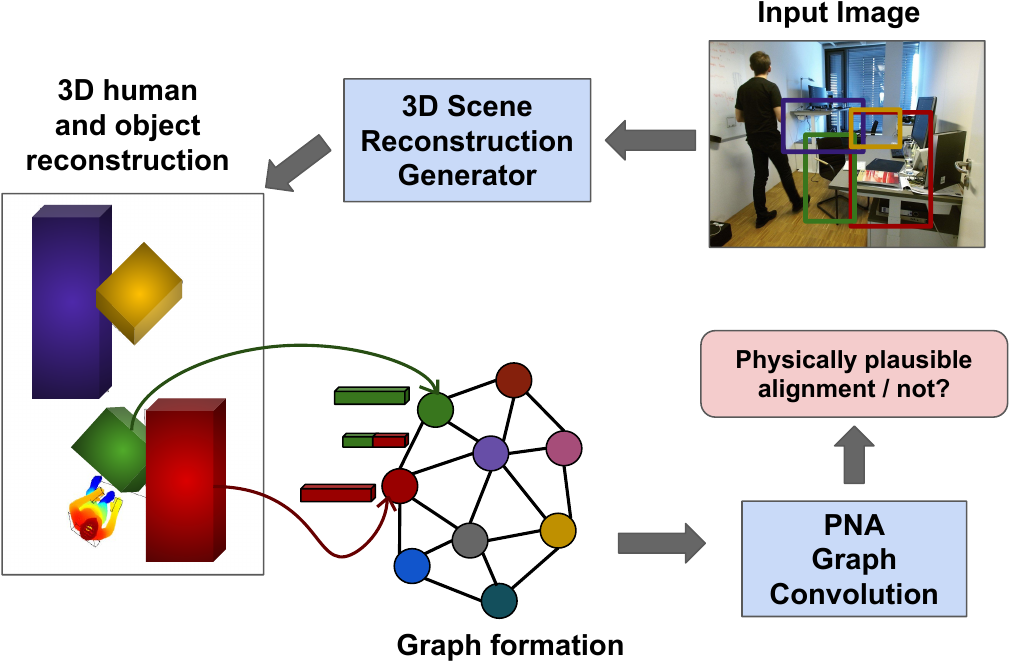}
    \caption{\footnotesize{Overview of our approach. From a monocular RGB image, we learn to generate physically plausible 3D human-scene reconstruction.
    We propose using adversarial training for analyzing the physical plausibility of the human and object alignments in the generated scene using implicit features from the 3D representations of the scene elements. Specifically, we form a graph using features extracted from the 3D bounding boxes around each scene element (Section \ref{sec:graph_disc}). We use a graph convolution network (PNAConv) to differentiate a plausible scene representation from an implausible one using the generated graph and to penalize the scene reconstruction system for physically implausible scene prediction.}}
    \label{fig:intro_fig}
\end{figure} 

The current optimization-based 3D human-scene reconstruction methods \cite{weng2021holistic,yi2022human} explicitly define some hand-crafted rules (contact and non-collision losses) to impose these physical laws and constraints between different scene elements. Explicitly encoding every physical law between different elements in the scene is difficult, as the applicability of these laws depends on the specific interactions between the elements, which may vary from frame to frame. For example, a person is moving a chair from one place to another and a person is sitting on a chair. \cite{weng2021holistic,yi2022human} employ contact and non-collision losses to ensure plausible reconstructions in such scenarios. Contact loss minimizes the distance between the human body mesh and the scene mesh, whereas, non-collision loss penalizes the penetration of human body mesh into the scene mesh or one object into another. However, for these two scenarios, such losses should be applied to different parts of the human body and the chair. Also, not only the contact or collision constraints but also the proper placement of the human on the chair is constrained by their relative orientations.
In this paper, we aim to learn these constraints implicitly from the training data for building a human-scene reconstruction system that can produce a physically plausible reconstruction without explicitly defining any physical laws and constraints.

During training, we propose representing the holistic scene as a graph and employing adversarial training with a graph discriminator to improve the realism of the reconstructions. Given a single RGB image, we utilize off-the-shelf networks \cite{kirillov2020pointrend,8765346} to predict 2D bounding boxes around human and object elements. Generator networks are then used to predict 3D reconstructions of these elements and form a graphical representation of the scene using the 3D bounding boxes around these 3D reconstructions. Nodes in the graph correspond to the human and object elements, while edges are defined by distance and angle features between pairs of elements (Fig. \ref{fig:intro_fig}). This helps comprehend the human-scene interactions and object-object relations in the scene for analyzing their relative positioning, orientations, and sizes from the training data. 
We propose using a graph neural network (GNN)-based discriminator \cite{corso2020principal} to differentiate between a graph formed from the \textit{predicted or reconstructed scene elements with implausible alignments} in 3D violating the physical laws/constraints from the \textit{real graph formed from the actual 3D scenes present in the training data} with physically plausible object alignments. If the predicted scene reconstruction does not match with the real distribution seen in the training data, it penalizes the human and object reconstruction generator to produce a physically plausible scene. Using a graph discriminator helps our method to learn the relations between different elements of the scene from the data implicitly, using different features that define the alignments of the elements in the scene.

Moreover, \cite{weng2021holistic,yi2022human} optimize the initial predictions from the off-the-shelf networks over multiple frames which is unsuitable for robotic applications in an online framework.
\textit{Our method does not require the whole sequence information a-priori, instead reconstructs the scene from a single monocular RGB image, resulting in much faster reconstruction compared to the existing methods.} In summary, our contributions are:
\begin{itemize}
    \item We propose representing a scene in the form of a graph using distance and angle features between each pair of elements in the scene that gives an implicit knowledge of the alignments of the scene elements.
    \item We propose an adversarial learning approach using a graph neural network-based discriminator for building a physical plausibility-aware 3D human-scene reconstruction system. 
    Using a GNN-based discriminator helps the generator to produce more accurate localizations and reconstructions of the scene elements.  
    \item Our single-image reconstruction method achieves comparable performance with the existing optimization-based methods that use multiple frames' information for reconstruction. Our learning-based model for holistic human-scene reconstruction is computationally more efficient compared to the existing methods.
\end{itemize}

\section{Related Work}

\noindent
\textbf{Scene-guided human pose and motion estimation:}
Recently, few methods have leveraged the scene information for accurate human reconstruction and motion predictions \cite{hassan2019resolving,rempe2021humor,shimada2020physcap,hassan2023synthesizing,luo2022embodied}. These methods optimize the human pose based on its contacts and collisions with the given ground-truth scene. Shimada \etal \cite{shimada2020physcap} use ground reaction forces to predict a temporally stable human pose. Hassan \etal \cite{hassan2023synthesizing} propose character animation with challenging scene interactions using an adversarial discriminator that assess the realism of the human motion in the context of the scene. Luo \etal \cite{luo2022embodied} propose human 3D pose estimation by utilizing motion imitation in prescanned 3D scenes. Few approaches have utilized the ground-truth scene information for accurate placement of the human \cite{hassan2021populating} or generating human \cite{PSI:2019} in the given scene. \textit{While previous works focus on enhancing human reconstruction and localization with ground-truth scene knowledge, our objective is to reconstruct both humans and the scene without any prior knowledge of the scene.}

\noindent
\textbf{Coarse scene reconstruction:}
Few techniques \cite{monszpart2019imapper,chen2019holistic++} focus on the coarse reconstruction of the scene. iMapper \cite{monszpart2019imapper} estimates object layout by leveraging human-object interactions inferred from 3D human skeleton data, while Chen \etal \cite{chen2019holistic++} utilize an HOI (human-object interactions) graph to learn spatial relations between the scene elements and enforce physical constraints. This method jointly infers 3D human skeleton, object bounding boxes, room layout, and camera pose from the image and scene graph. \textit{These methods only generate coarse reconstruction and do not produce the detailed shape of the scene objects and humans.} Some recent approaches \cite{zhang2020perceiving,xie2022chore,dabral2021gravity} partially reconstruct the scene, restricting their scope to only those objects which are in contact with the human. Zhang \etal \cite{zhang2020perceiving} use prior knowledge of the object size distribution to infer the intrinsic scale of the objects. Dabral \etal \cite{dabral2021gravity} propose a gravity-aware system for reconstructing 3D human and free-flight objects by joint optimization of the human and object trajectories. Xie \etal \cite{xie2022chore} reconstruct the scene via an implicit representation (unsigned distance fields) of the humans and objects. 
These methods do not provide a holistic reconstruction of the scene and focus only on a single object that is in contact with the human. \textit{In contrast, our aim is to generate a detailed mesh representation of the entire scene (or multiple objects).}

\noindent
\textbf{Holistic scene reconstruction:} Weng \etal \cite{weng2021holistic} first propose a holistic scene reconstruction. In this work, the authors jointly optimize the human mesh and object reconstruction network parameters based on explicitly defined physical constraints on surface contacts and interpenetration without any precise knowledge about the HSI (Human-Scene Interactions). Yi \etal \cite{yi2022human} incorporate explicit human scene interactions using POSA (Pose with prOximitieS and contActs) \cite{hassan2021populating}. POSA predicts the probable contact regions of a human body for a given human pose and scene. It also predicts the probable scene elements with which the human surface may come into contact. Yi \etal use this information for explicitly incorporating human-scene interactions as constraints during optimization over a sequence of frames for final scene reconstruction. Moreover, both these methods \cite{weng2021holistic,yi2022human} use optimization at the inference time, which is time-consuming. \cite{weng2021holistic} performs 4 stages of optimization, 1) initial human prediction, and optimization of the scene prediction network using 2) within-scene losses 3) body-scene losses, and 4) final optimization of human mesh based on the optimized scene. Moreover, they optimize the network parameters over the whole sequence for the final reconstruction. Whereas, \cite{yi2022human} does not optimize the network parameters but the initial predictions over a sequence. These methods rely on multiple frames for a reasonable reconstruction of the scene. \textit{However, our goal is to build a learning-based model that learns the physical laws and constraints implicitly from the training data and produces per-frame plausible reconstruction without inference-time optimization. This makes our model more time efficient and applicable for robotic applications in an online framework.}

\section{Method}

\begin{figure*}[ht]
\centering
\begin{subfigure}[b]{0.49\textwidth}
         \centering
         \includegraphics[width=0.99\columnwidth]{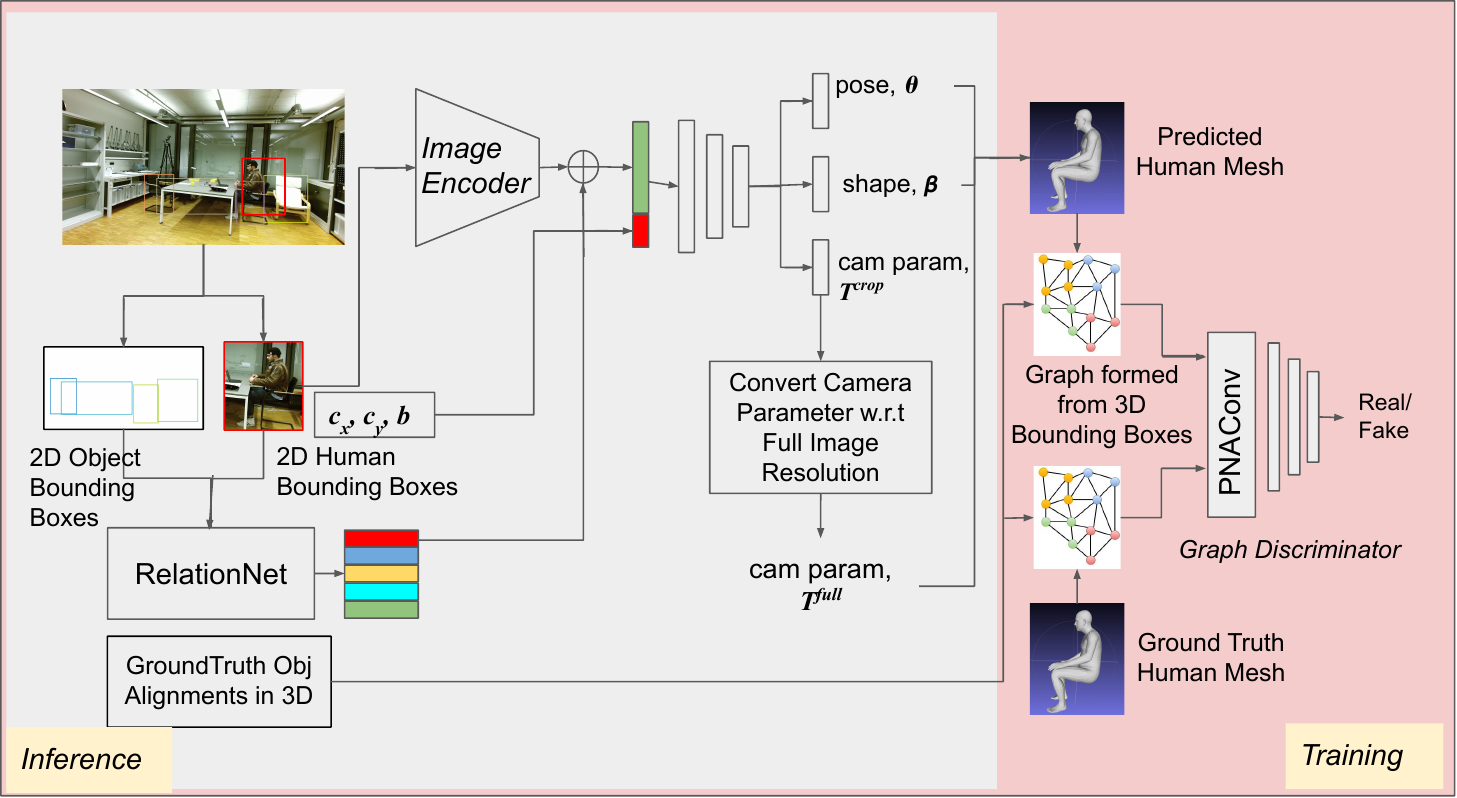}
         \caption{}
         \label{fig:bd_human}
     \end{subfigure}
     \begin{subfigure}[b]{0.49\textwidth}
         \centering
         \includegraphics[width=0.94\columnwidth]{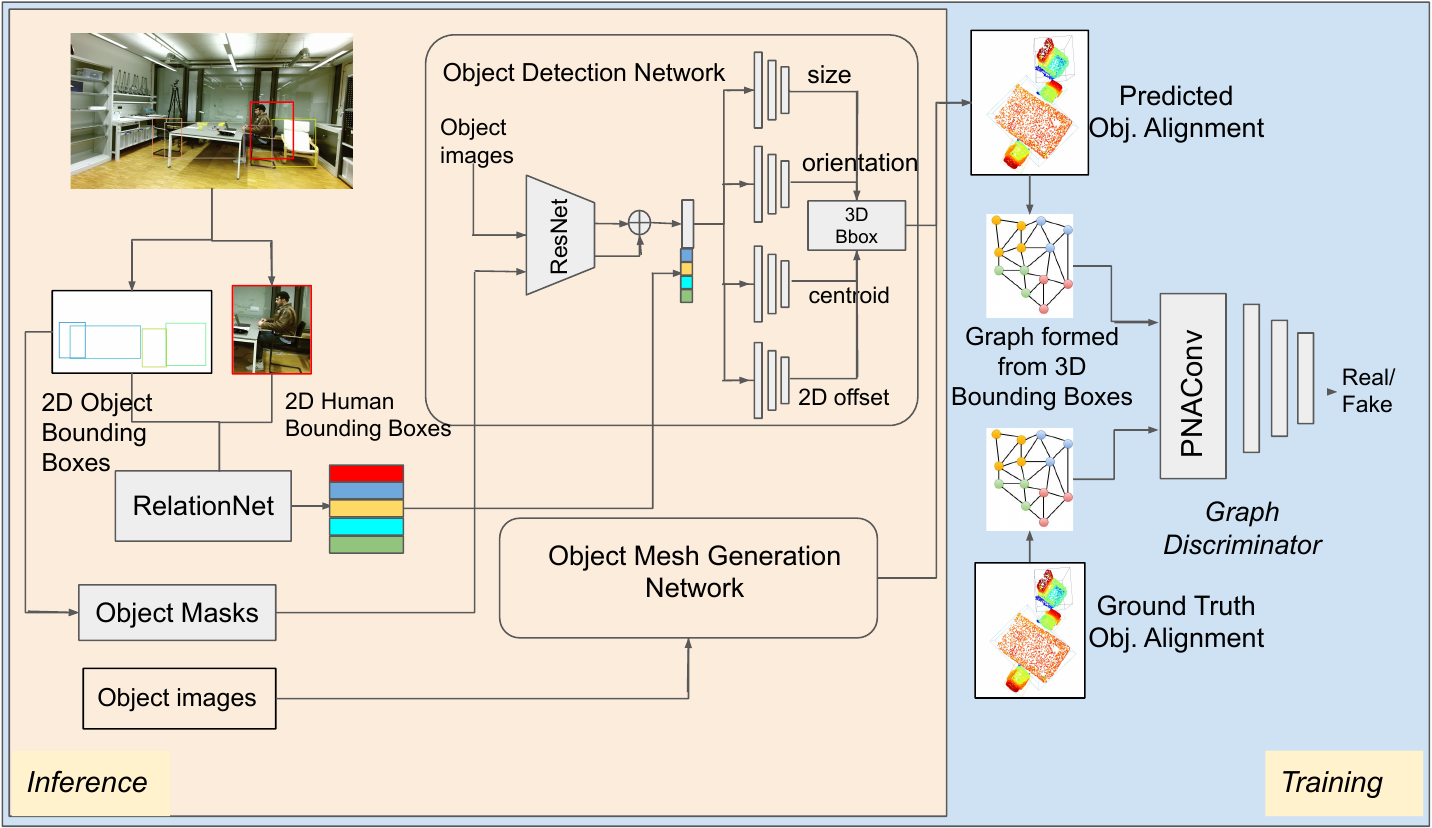}         
         \caption{}
         \label{fig:bd_object}
     \end{subfigure}     
         \caption{\footnotesize{(a) \textit{Human reconstruction module}: At inference time, given an input image, using information about bounding box around human (size, $b$ and relative distance of bounding box center \wrt image center, $(c_x, c_y)$), and relation feature of human with respect to the other objects in the scene ($\mathcal{F}_{h-obj}$), the human reconstruction network predicts the SMPL pose and shape parameters along with location information of the human, $T^{full}$  within the scene. (b) \textit{Object reconstruction module}: At inference time, given an input image, using object binary masks, and relation feature with respect to the other objects in the scene ($\mathcal{F}_{obj-obj}$), the object detection network predicts the object orientation, size, and centroid. Object mesh generation network weights are initialized from Total3D \cite{Nie_2020_CVPR} proposed mesh generation network and have not been updated in the training process.}}
\end{figure*}

\begin{figure*}[ht]
\centering
\includegraphics[width=0.75\textwidth]{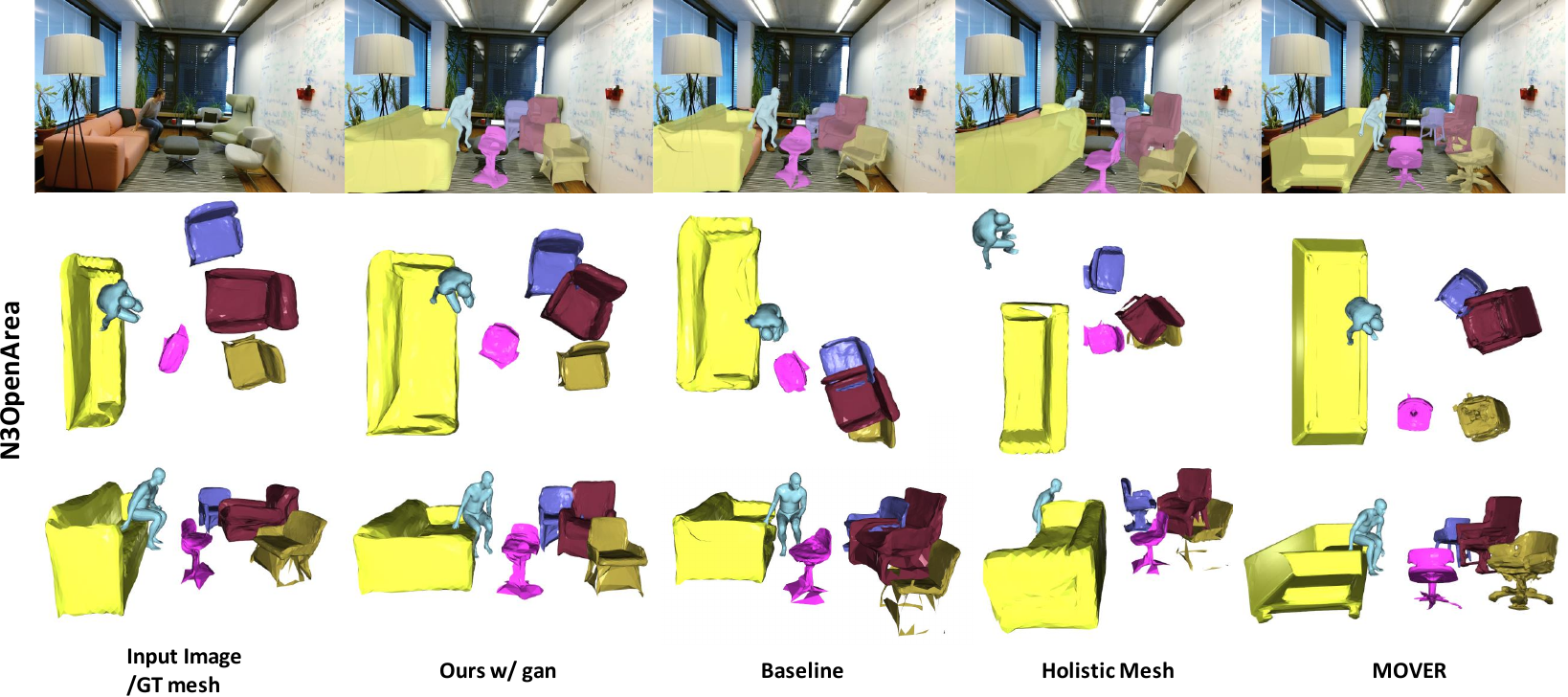}
\includegraphics[width=0.75\textwidth]{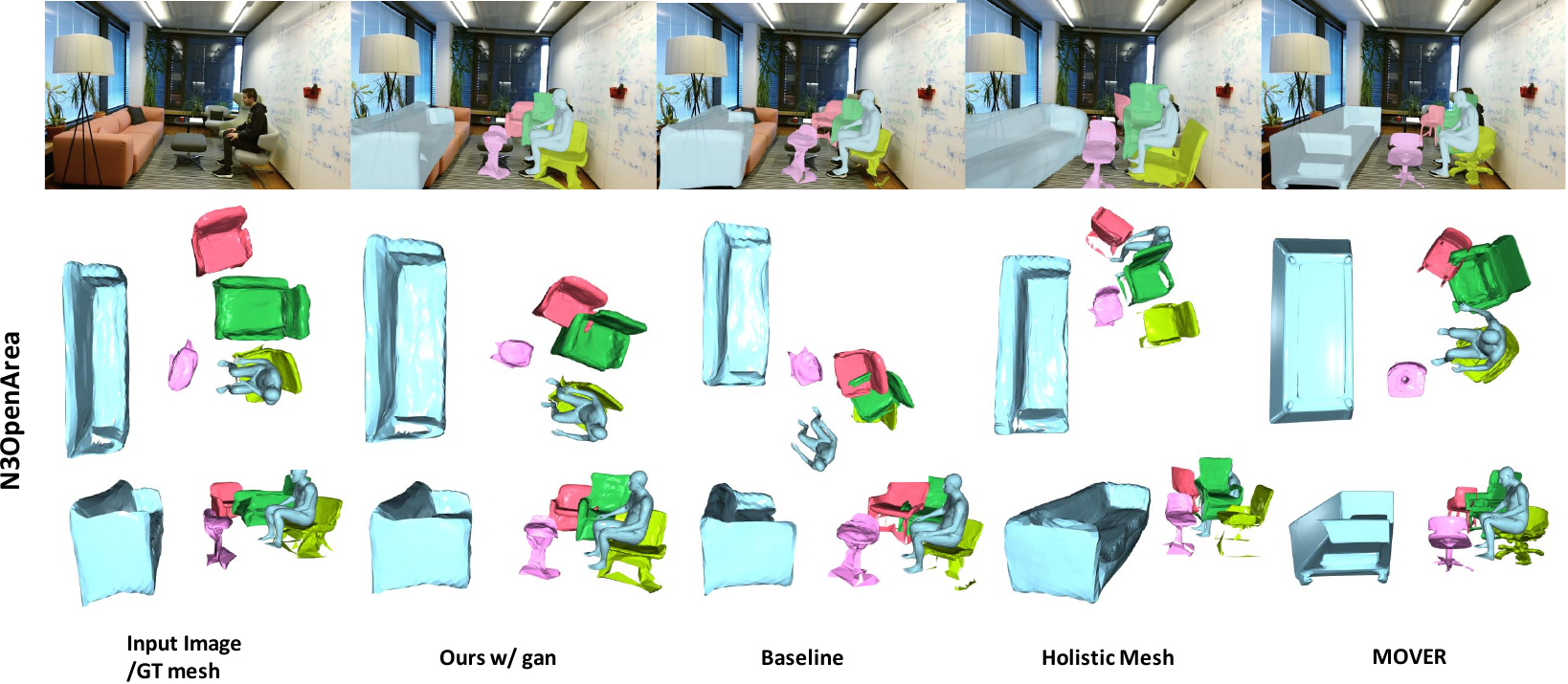}
\includegraphics[width=0.75\textwidth]{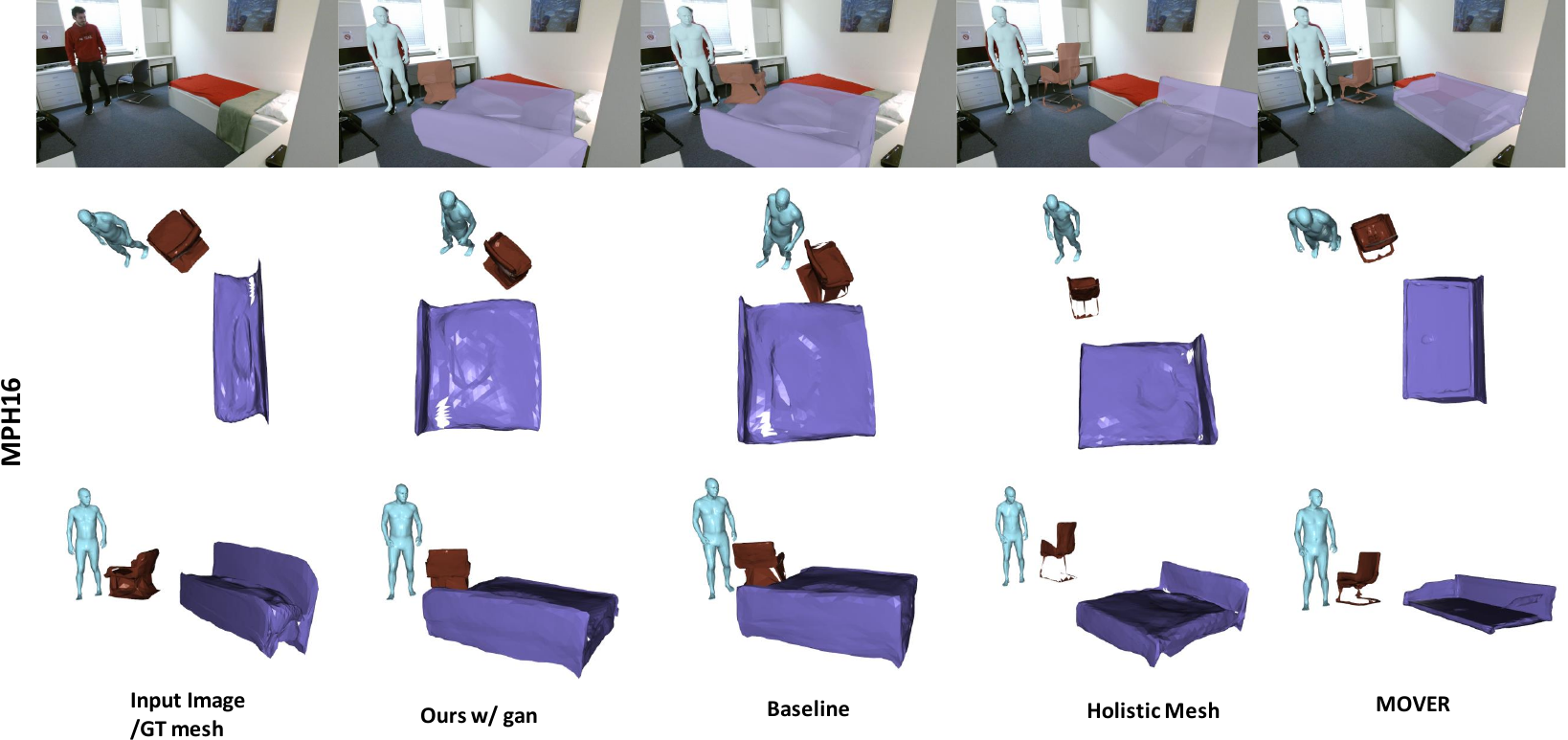}
\caption{\footnotesize{Scene reconstruction results on 2 scenes (N3OpenArea and MPH16) of PROX qualitative dataset. For every example, the second and third rows show reconstruction results from the top and side views. For a fair comparison with our single image-based method, results of HolisticMesh \cite{weng2021holistic} are generated by optimizing their network on a single image. For MOVER \cite{yi2022human}, results are rendered from  reconstructed results shared on MOVER's website.}}
\label{fig:prox_qual}
\end{figure*}

Given an indoor image, we use pre-trained networks to get 2D bounding boxes around objects and humans and generate 3D reconstructions. The predicted 3D reconstructions are used to form a graphical representation of the scene and analyzed using a graph discriminator compared to the graph created from the ground truth. This section explains our 1) human mesh reconstruction (Fig. \ref{fig:bd_human}), 2) object reconstruction (Fig. \ref{fig:bd_object}), and 3) the graph discriminator modules.

\subsection{Human mesh reconstruction:} 
We use a learning-based model \cite{li2022cliff} for human body reconstruction. Given an image firstly we use OpenPose \cite{8765346} network for human bounding box detection. The human reconstruction network takes the cropped human image and bounding box information \ie location of the bounding box center ($c_x, c_y$) relative to the actual image center ($W/2, H/2$), and the size of the squared bounding box ($b$) in actual image scale (before cropping and resizing) as input to predict the SMPL (\textbf{S}kinned \textbf{M}ulti-\textbf{P}erson \textbf{L}inear model) \cite{loper2015smpl} body parameters ($\theta \in \mathbb{R}^{72}$, orientation of each body joint \wrt the root \ie pelvis joint and $\beta \in \mathbb{R}^{10}$, representing body shape) and the root joint translations, $T^{full} = [t_X^{full},t_Y^{full},t_Z^{full}]$, relative to the original camera.

\noindent
\textbf{Object-aware human reconstruction:} 
A human's pose and location are conditioned by the other objects' positions in the scene; hence it is necessary to incorporate the positional relation of the human with the other scene elements. Nie \etal \cite{Nie_2020_CVPR} propose using the relative positions of surrounding objects in the scene for each object bounding box prediction. Following a similar approach, we use RelationNet \cite{hu2018relation} to extract the relation features of the human, $\mathcal{F}_{h-obj} \in \mathbb{R}^{2048}$, with respect to the other elements in the scene. The inputs to RelationNet \cite{hu2018relation} are i) the ResNet-34 feature for each of the scene elements, and ii) geometric features ($ \in \mathbf{R}^{n \times 64}$) between each pair of scene elements ($n$) defined by the relative distance between the 2D bounding boxes. We add this relation feature with the image feature $\mathcal{F}_{img} \in \mathbb{R}^{2048}$ of the cropped human instance. 

The combined bounding box feature and image-object feature $\mathcal{F}_{bbox} \oplus (\mathcal{F}_{img} + \mathcal{F}_{h-obj})$ is passed through an MLP regressor proposed in \cite{li2022cliff} to predict SMPL pose ($\theta$), shape ($\beta$) parameters and the perspective camera translation $t^{crop} = [t^{crop}_X,t^{crop}_Y,t^{crop}_Z]$ with respect to the cropped image. This perspective camera translation is then modified to get the body root translation with respect to the original camera,  $T^{full}$ using the focal length $f$ and the human bounding box information $(c_x, c_y, b)$ \cite{li2022cliff}.
$T^{full}$ is added to the SMPL body vertices to get the human body reconstruction in the world coordinates. From the SMPL body vertices, 3D body joints are predicted using the joint regressor \cite{kolotouros2019learning}. The losses used for training the human body generator are as follows,

\begin{equation}
\begin{split}
\mathcal{L}_{human} & = \lambda_{pose}\mathcal{L}_{pose} +  \lambda_{shape}\mathcal{L}_{shape} + \lambda_{2D}\mathcal{L}_{2D} \\ +  \lambda_{3D}&\mathcal{L}_{3D} + \lambda_{verts}\mathcal{L}_{verts} + \lambda_{h-obj_{gan}}\mathcal{L}_{h-obj_{gan}}
\end{split}
\label{equ:h_recon_loss}
\end{equation}

Where, $\mathcal{L}_{pose} = ||\theta - \theta^\prime||_2$, $\mathcal{L}_{shape}=|\beta - \beta^\prime|$, $\mathcal{L}_{3D}=||J_{3D} - J_{3D}^\prime||_2$, $\mathcal{L}_{2D}=||J_{2D} - J_{2D}^\prime||_2$, and $\mathcal{L}_{verts}=||\mathcal{B}_{verts} - \mathcal{B}_{verts}^\prime||_2$ are losses applied on predicted SMPL pose, shape parameter, 3D body joints, reprojected 3D joints, and body mesh vertices $\{\theta^\prime,\beta^\prime, J_{3D}^\prime, J_{2D}^\prime,\mathcal{B}_{verts}\}$. The adversarial loss is defined as follows,
\begin{equation}
\begin{split}
    \mathcal L_{h-obj_{gan}}  = \mathbb{E}_{\{\theta^\prime,\beta^\prime,\mathcal{B}_v^\prime\}} & [log(D(\mathcal{G}_h^\prime)] \\ & + \mathbb{E}_{\{\theta,\beta,\mathcal{B}_v\}}[log (1-D(\mathcal{G})] 
    \end{split}
\end{equation} 

where $\mathcal{G}_h^\prime$ is the graph formed from predicted human mesh and ground-truth object meshes and $\mathcal{G}$ is formed from ground-truth human mesh and object meshes. Details about graph formation are described in Section \ref{sec:graph_disc}.

\subsection{Object reconstruction:}
\label{sec:obj_recon}
Similar to \cite{weng2021holistic} and \cite{yi2022human}, we use the object reconstruction module proposed by Nie \etal\cite{Nie_2020_CVPR}. It consists of three sub-networks, i.e., Layout Estimation Network (LEN), Object Detection Network (ODN) for object bounding box prediction, and Mesh Generation Network (MGN). From a given image, first object labels and 2D bounding boxes are extracted using an off-the-shelf network PointRend \cite{kirillov2020pointrend}. Then ResNet-34 \cite{he2016deep} is used to get the image feature from the object patches and RelationNet \cite{hu2018relation} is used to extract the object-object relation feature $\mathcal{F}_{obj-obj}$. Using this feature the Object Detection Network predicts each object's 
size ($\mathbb{B}_{s}^\prime$), orientation ($\mathbb{B}_{\theta}^\prime$), and 2D offset of the bounding box center. We use ground truth camera parameters to unproject the predicted 2D object center for computing the 3D object centroid ($\mathbb{B}_{c}^\prime$). The pre-trained Mesh Generation Network (MGN) from \cite{Nie_2020_CVPR} is used to get the object meshes, which are then resized, rotated, and translated using the predicted object orientation ($\mathbb{B}_{\theta}^\prime$), centroid ($\mathbb{B}_{c}^\prime$) and size ($\mathbb{B}_{s}^\prime$) information. We use the same losses as proposed in \cite{Nie_2020_CVPR} for training.

As we perform per-frame reconstruction human occlusion highly affects the object reconstruction. Hence, to further enhance the object detection performance for occlusion cases, we propose using occlusion masks' features along with the ResNet image features (Fig. \ref{fig:bd_object}). For training purposes, we use ground-truth occlusion masks. Ground-truth occlusion masks are extracted by using $ M_{obj} - M_{obj} \cap M_{human}$, where $M_{obj}$ are unoccluded object masks that are extracted manually from a keyframe (without occlusion by the human) and $M_{human}$ is the rendered body mask. 
The losses used for training the human body generator are as follows,

\begin{equation}
\begin{split}
\mathcal{L}_{obj} & = \lambda_{size}\mathcal{L}_{size} +  \lambda_{ori}\mathcal{L}_{ori}\\ & + \lambda_{centroid}\mathcal{L}_{centroid} + \lambda_{obj-obj_{gan}}\mathcal{L}_{obj-obj_{gan}}
\end{split}
\label{equ:obj_recon_loss}
\end{equation}
Where, $\mathcal{L}_{size} = ||\mathbb{B}_s - \mathbb{B}_s^\prime||_2$, $\mathcal{L}_{ori}=||\mathbb{B}_{\theta} - \mathbb{B}_{\theta}^\prime||_2$, and $\mathcal{L}_{centroid}=||\mathbb{B}_c - \mathbb{B}_c^\prime||_2$ are losses applied on predicted size, orientation and centroid of the object bounding boxes $\{\mathbb{B}_{s}^\prime,\mathbb{B}_{\theta}^\prime,\mathbb{B}_{c}^\prime\}$. The adversarial loss is defined as,

\begin{equation}
\begin{split}
    \mathcal L_{obj-obj_{gan}} = \mathbb{E}_{\{\mathbb{B}_{s}^\prime,\mathbb{B}_{\theta}^\prime,\mathbb{B}_{c}^\prime\}} & [log(D(\mathcal{G}_{obj}^\prime)] + \\ & \mathbb{E}_{\{\mathbb{B}_{s},\mathbb{B}_{\theta},\mathbb{B}_{c}\}}[log (1-D(\mathcal{G})] 
    \end{split}
\end{equation} 
where $\mathcal{G}_{obj}^\prime$ and $\mathcal{G}$ are the graphs formed from predicted and ground-truth object meshes.

\subsection{Graph Discriminator:}
\label{sec:graph_disc}
We propose using the GNN \cite{corso2020principal} for analyzing the physical laws and constraints using implicit feature representation of the reconstructed 3D scene elements. Specifically, we use a graph discriminator to distinguish a physically plausible real scene alignment from a reconstructed physically implausible scene to further strengthen the performance of the object and human reconstruction modules.

\noindent
\textbf{Graph formation:}
We form a graph, $\mathcal{G}(\mathcal{V},\mathcal{E})$, $\mathcal{V} \in \mathbb{R}^{P}, P = N+M$ from the scene, where, nodes $\mathcal{V}$, are represented by the objects $\{O_n\}_{n=1,2,...N}$ and human body segments $\{H_m\}_{m=1,2,...M}$, that come in contact (\eg feet while walking or standing on the floor, hands while touching any objects \etc) with the scene or scene objects. These body segments, $\{H_m\}_{m=1,2,...M}$ are defined by the vertices of the corresponding regions of the body mesh as proposed by Hassan \etal in \cite{hassan2019resolving}. $N$ is the number of objects in the scene and $M$ is the number of body segments under consideration. We consider the corner points of the 3D bounding box around the objects or the human body segments as the node features $\mathcal{F}_{\mathcal{V}_l} \in \mathbb{R}^{8 \times 3}$. Implementation details for 3D bounding box formation around human body segments are described in Section \ref{sec:impl}. The edge features $\mathcal{F}_{\mathcal{V}_l -> \mathcal{V}_k} = \mathcal{F}_{\mathcal{V}_l,\mathcal{V}_k}^{dist} \oplus \mathcal{F}_{\mathcal{V}_l,\mathcal{V}_k}^{surf-norm}$ are defined by the Euclidean distance $\mathcal{F}_{\mathcal{V}_l,\mathcal{V}_k}^{dist}$ between every pair of corners (Eqn. \ref{eqn:vert_dist}) and surface normals $\mathcal{F}_{\mathcal{V}_l,\mathcal{V}_k}^{surf-norm}$ of every pair of faces (Eqn. \ref{eqn:face_norm}) of the bounding boxes, between each pair of nodes or scene elements $\mathcal{V}_l$ and $\mathcal{V}_k$. 
$\mathcal{F}_{\mathcal{V}_l,\mathcal{V}_k}^{dist}$ helps to analyze the relative distance, positioning, and sizes of the interacting elements whereas, $\mathcal{F}_{\mathcal{V}_l,\mathcal{V}_k}^{surf-norm}$ helps to analyze the relative orientations.

\begin{equation}
\mathcal{F}_{\mathcal{V}^i_l, \mathcal{V}^j_k}^{dist} = \lVert \mathbb{B}^l_i - \mathbb{B}^k_j \rVert, \; 
\label{eqn:vert_dist}
\end{equation}

where, $\mathbb{B}^l_i,\mathbb{B}^k_j \in \mathbb{R}^{8 \times 1}$ represents the $i^{th}$ and $j^{th}$ corner points of the 3D bounding boxes of $l^{th}$- and $k^{th}$-object.

\begin{equation}
\mathcal{F}_{\mathcal{V}^i_l, \mathcal{V}^j_k}^{surf-norm} = \mathcal{B}^l_{n_i} \times \mathcal{B}^k_{n_j}, \; 
\label{eqn:face_norm}
\end{equation}

where, $\mathbb{B}^l_{n_i},\mathcal{B}^k_{n_j} \in \mathcal{R}^{6 \times 3}$ represents the surface normals on $i^{th}$ and $j^{th}$ faces of $l^{th}$- and $k^{th}$-object bounding boxes. More details about the network architecture and training are given in Section \ref{sec:impl}. 

\noindent
\textbf{Scene representation learning:} We have used Principal Neighbourhood Aggregation (PNA) graph convolution \cite{corso2020principal} network for scene representation learning. Unlike the traditional graph convolution networks, PNA use multiple aggregators 
for effective feature representation learning, as a single message aggregator fails to capture meaningful representations from different kinds of messages \cite{corso2020principal}. Also, it uses a scaler to either amplify or attenuate the aggregated message at a node, based on the number of messages coming to the node. We have used four aggregators i.e \textit{mean}, \textit{max}, \textit{min} and \textit{std}  and three-scalers i.e. \textit{identity}, \textit{amplification} ($\alpha=1$), \textit{attenuation} ($\alpha=-1$) for effective feature aggregation from neighborhood nodes. 
Node embeddings learned using PNAConv layers are passed through a graph pooling layer to get the scene representation. Finally, an MLP is used with an output layer to classify the graph between real and predicted. Hence, the output of this graph discriminator consists of binary labels: 0 for predicted and 1 for ground truth or real.

\section{Experiments and Results}
\subsection{Dataset}

We use the PROX (Proximal Relationships
with Object eXclusion) dataset \cite{hassan2019resolving} for our experiments. It consists of a) \textit{qualitative} and b) \textit{quantitative} datasets. \textit{PROX qualitative} dataset contains 12 different scenes including, bedrooms, libraries, offices, and living rooms. It contains around 100K RGB-D frames with humans interacting with the scenes. 
It contains pseudo-ground truth for the body poses generated from the SMPLify-X \cite{pavlakos2019expressive} method using depth and RGB frames. The \textit{PROX quantitative} contains a single scene where human activities or interactions with the scene have been captured using MoCap. The ground truth of human body mesh is generated using MoSh++ \cite{mahmood2019amass}. The scene is scanned and reconstructed using Structure Sensor and Skanect. This dataset contains a total of 18 sequences with 180 RGB-D frames. 
We have used this dataset for cross-dataset evaluation. 
We use \textit{PROX qualitative} and perform data augmentation (scaling and translation) for the training of both human and object reconstruction modules. We use a similar train and test split as defined by Hassan \etal in POSA \cite{hassan2021populating}. Amongst 12 scenes, 7 scenes have been used for training and 3 scenes (N3OpenArea, MPH1Library, and MPH16) have been used for testing. We also provide qualitative results for human and scene reconstruction on \textit{PROX quantitative} dataset.

\subsection{Implementation Details}
\label{sec:impl}
We use the CLIFF (\textbf{C}arrying \textbf{L}ocation \textbf{I}nformation in \textbf{F}ull \textbf{F}rames) model proposed by Li \etal \cite{li2022cliff} as the backbone of our human mesh reconstruction module and train this module on the \textit{PROX qualitative} dataset. 
We additionally use object alignment information in 2D as input for more accurate localization of the human in the scene (Fig. \ref{fig:bd_human}).
For object reconstruction, we use the network proposed by \cite{Nie_2020_CVPR} as our backbone network and initialized the weights of our object reconstruction network using the model proposed in \cite{Nie_2020_CVPR}. 
We additionally use object occlusion masks as input (Fig. \ref{fig:bd_object}).
For creating a graph from the predicted scene elements, we calculate 3D bounding boxes around human body segments that come in contact with the scene, \eg feet, hands \etc. \\
\textbf{3D bounding box creation around human mesh:}
We perform PCA \cite{pcaref} on the projection of vertices of each body segment on each plane for calculating the orientation around the perpendicular axis and size of the bounding boxes. For example, to calculate the orientation of a body segment around the $x$-axis, the body segment vertices are projected on the $yz$ plane ($x=0$). PCA is performed on the projected 2d points (on $yz$ plane) to calculate the principal orientation ($\theta_x$) of the point cluster. Using the calculated orientation, the points are aligned with the axis (applying $-\theta_x$) to calculate the size of the body segment. The orientations around the $y$- and $z$-axis are calculated in a similar manner. Finally, the 3D bounding box for each body segment vertices is formed using these size, orientation, and centroid information for graph formation.

Our proposed graph discriminator consists of 4 PNAConv layers \cite{corso2020principal} followed by a graph pooling layer, \textit{global$\_$add$\_$pool}, and 3 fully connected layers with ReLU at the intermediate and Sigmoid activation at the last layer. The \textit{global$\_$add$\_$pool} layer calculates a graph level output \ie a representation of the input graph, by adding the node features across the node dimension of the input graph. We have used \textit{global$\_$add$\_$pool} layer from pytorch\_geometric library \cite{FeyLenssen2019}.

We train human and object reconstruction modules independently. We use Adam optimizer \cite{kingma2014adam} with the initial learning rate of $1e-4$ and $1e-3$ for the object and human reconstruction network training. We have used PyTorch and PyTorch geometric library for implementation and GTX 1080i GPU for training and inference. In the human reconstruction training loss (Eqn. ~\ref{equ:h_recon_loss}) $\lambda_{verts} = 1.0 $, $\lambda_{pose} = 1.0 $, $\lambda_{shape} = 0.1 $, $\lambda_{2D} = 5.0 $ and $\lambda_{3D} = 5.0$. The total loss is scaled by 60. In the object reconstruction loss (Eqn. \ref{equ:obj_recon_loss}) are $\lambda_{ori} = 1.0 $, $\lambda_{size} = 1.0 $, $\lambda_{centroid} = 1.0 $, $\lambda_{gen} = 0.1 $ and $\lambda_{disc} = 0.1$. The total loss is scaled by 100. 

\begin{table}[ht]
\centering
    \noindent\resizebox{\columnwidth}{!}{
\begin{tabular}{|c|c|c|c|c|c|}
\hline 
Method & Input & \multicolumn{2}{c|}{Object Reconstruction Quality} & \multicolumn{2}{c|}{Physical Metrics} \\
\hline 
\vspace{1pt}
 &  & IoU3D $\uparrow$ & IoU2D $\uparrow *$  & Non-collision $\uparrow$  & Contact $\uparrow$ \\
\hline
MOVER \cite{yi2022human} & Image Sequence & \textcolor{red}{0.3665} & \textcolor{red}{0.6179} & 0.9992 & 0.1601 \\ 
HolisticMesh \cite{weng2021holistic} & Image Sequence & 0.2607 & 0.4361 & 0.8878 & 0.7725 \\ 
\hline
\hline
Total3D \cite{Nie_2020_CVPR} & Single Image & 0.1917 & 0.3038 & - & - \\
Baseline $\dagger$  & Single Image & 0.3156 & 0.4037 & - & - \\
Ours & Single Image & \textcolor{blue}{0.3470} & \textcolor{blue}{0.5515} & \textcolor{blue}{0.9011} & \textcolor{blue}{0.7897} 
 \\
\hline
\end{tabular}
}
\caption{\footnotesize{Quantitative comparison of our object reconstruction method with SoTA on \textit{PROX qualitative} dataset. Blue denotes the best-performing method that uses single image input and performs per-frame prediction. Red denotes the overall best-performing method. $*$ IoU2D is calculated as bird's eye view 2D bounding box IoU. $\dagger$ Baseline is Total3D \cite{Nie_2020_CVPR} trained on \textit{PROX qualitative} dataset.}}
\label{tab:quantitative}
\end{table}

\begin{table}[ht]
\centering
    \noindent\resizebox{0.7\columnwidth}{!}{
\begin{tabular}{|c|c|c|}
\hline 
 & \multicolumn{2}{c|}{Object reconstruction} \\
\hline  
 Method &  IoU3D $\uparrow$ & IoU2D $\uparrow$ \\
\hline
HolisticMesh \cite{weng2021holistic} & 0.2048 & 0.5167 \\
Baseline + object mask & 0.2514 & 0.5208 \\
Baseline + gan + object mask (Ours) & 0.2647 & 0.6097 \\
\hline
\end{tabular}
}
\caption{\footnotesize{Quantitative comparison of our object reconstruction method with SoTA on \textit{PROX qualitative} dataset using predicted 2D bounding box and occlusion masks. The model trained with graph discriminator is defined as 'gan'.}}
\label{tab:quantitative_predbbox}
\end{table}

\begin{table}[ht]
\centering
    \noindent\resizebox{\columnwidth}{!}{
\begin{tabular}{|c|c|c|c|c|c|}
\hline 
Dataset & Method & \multicolumn{2}{c|}{Localization} & \multicolumn{2}{c|}{Pose Estimation} \\
\hline 
 &  & MPJPE & V2V & p-MPJPE & p-V2V \\
\hline
\multirow{2}{*}{PROX quantitative} & MOVER \cite{yi2022human} & \textbf{174.37} & \textbf{178.31} & 73.60 & 67.89 \\
& HolisticMesh \cite{weng2021holistic} & 190.78 & 192.21 & 72.72 & \textbf{61.01} \\
& Ours $\dagger$ & 266.31 & 268.08 & \textbf{72.31} & 84.14 \\
\hline
\hline
\multirow{2}{*}{\makecell{PROX quantitative \\ (Testset)}} & HolisticMesh \cite{weng2021holistic} & 187.24 & 191.00 & 65.13 & \textbf{64.07} \\
& Ours $\ddagger$ & \textbf{181.02} & \textbf{187.89} & \textbf{55.82} & 67.62
 \\
\hline
\hline
PROX qualitative & Ours $\dagger$ & 182.57 & 187.79 & 59.51 & 70.14 \\
\hline
\end{tabular}
}
\caption{\footnotesize{Quantitative comparison of our human reconstruction with SoTA. Ours $\dagger$: Human reconstruction model trained on \textit{PROX qualitative} dataset; Ours $\ddagger$: Human reconstruction model fine-tuned on \textit{PROX quantitative} dataset using 5-fold cross-validation method.}}
\label{tab:quantitative_human}
\end{table}

\begin{table}[ht]
\centering
    \noindent\resizebox{0.7\columnwidth}{!}{
\begin{tabular}{|c|c|c|}
\hline 
 & \multicolumn{2}{c|}{Object reconstruction} \\
\hline  
 Method &  IoU3D $\uparrow$ & IoU2D $\uparrow$ \\
\hline
Baseline & 0.3156 & 0.4037 \\
Baseline + gan & 0.3279 & 0.4206 \\
Baseline + object mask & 0.3059 & 0.5208 \\
Baseline + gan + object mask (Ours) & \textbf{0.3470} & \textbf{0.5515} \\
\hline
\end{tabular}
}
\caption{\footnotesize{Ablation for object reconstruction on \textit{PROX qualitative}.}}
\label{tab:quantitative_abl}
\end{table}

\subsection{Quantitative Results:}
\noindent
\textbf{Object reconstruction}
We have evaluated and compared our object reconstruction results against the state-of-the-art methods HolisticMesh \cite{weng2021holistic} and MOVER \cite{yi2022human} in Table \ref{tab:quantitative} (with GT 2D bounding box and occlusion masks) and \ref{tab:quantitative_predbbox} (with predicted 2D bounding box and occlusion masks by PointRend \cite{kirillov2020pointrend}). It should be noted that MOVER optimizes the object and human reconstruction based on the HSI information collected from a sequence of frames taken as input. And HolisticMesh performs optimization of the object reconstruction network parameters over the sequence of frames. We have used the same metrics proposed in \cite{weng2021holistic,yi2022human} for the performance evaluation. As we use \textit{PROX qualitative} dataset for training, the performance metrics are evaluated on 3 test scenes of the dataset for all three methods.
Similar to \cite{weng2021holistic,yi2022human,Nie_2020_CVPR} we have used 3D and 2D IoU for quantifying the object localization accuracy. 
We train the Total3D model \cite{Nie_2020_CVPR} on PROX qualitative dataset and consider it as our \textit{Baseline} model.
Our method \ie \textit{Baseline} with graph discriminator outperforms the results of \textit{Baseline} and HolisticMesh in 3D IoU using both predicted (Table \ref{tab:quantitative_predbbox}) and ground-truth (Table \ref{tab:quantitative}) bounding box and occlusion masks.
Following MOVER \cite{yi2022human}, we have computed the Non-collision and Contact scores to assess the physical plausibility of the reconstructed scenes with respect to the reconstructed human (Table \ref{tab:quantitative}).
Total3D \cite{Nie_2020_CVPR} performs only object reconstruction, hence does not have non-collision and contact scores. 

\begin{figure}[ht]
\centering
 \includegraphics[width=0.75\columnwidth]{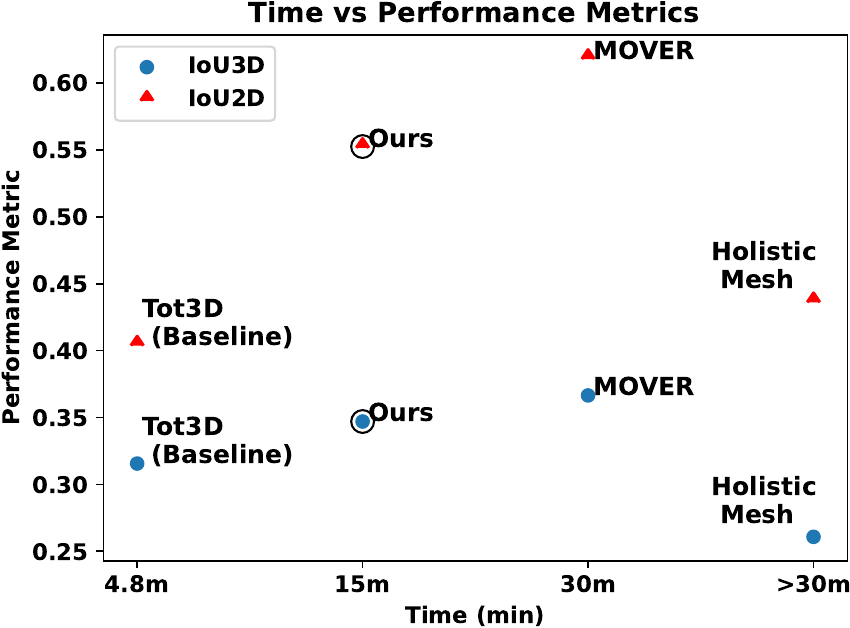}
    \caption{\footnotesize{Comparison of execution time vs. performance of our scene reconstruction method with respect to state-of-the-art methods. }} 
    \label{fig:inference_time}
\end{figure}

\noindent
\textbf{Human reconstruction}
Following the state-of-the-art methods we use Mean Per Joint Error (MPJPE) and vertex-to-vertex (V2V) distance for assessing the quality of human reconstruction (Table \ref{tab:quantitative_human}) \cite{kolotouros2019learning}. MPJPE is calculated on the 3D body skeleton joints without making the root position zero.
Our human reconstruction module is trained on human body meshes from 7 scenes of the \textit{PROX qualitative} dataset. 
Our method gives a reasonable performance on cross-dataset (\textit{PROX quantitative}) compared to the state-of-the-art methods. We also fine-tune the human reconstruction module using 12 sequences of PROX-quantitative data, which performs better than that of HolisticMesh \cite{weng2021holistic}. 
We use p-MPJPE and p-V2V for estimating the quality of human pose reconstruction similar to the state-of-the-art methods. 
Our method achieves the best p-MPJPE compared to the state-of-the-art methods, hence the human reconstruction module can capture the body pose correctly.

\subsection{Qualitative Results:}

We demonstrate our scene reconstruction results on two unseen scenes from the PROX qualitative dataset (Fig. \ref{fig:prox_qual}). For each frame, the reprojection of reconstructed object meshes on the image plane is shown in the first row, and 3D reconstructions from different viewing angles are in the second and third rows. Our approach, with the discriminator network (second column, Fig. \ref{fig:prox_qual}), produces physically plausible scene reconstructions with accurate relative positioning of humans and objects compared to the backbone model (without discriminator network, third column, Fig. \ref{fig:prox_qual}). However, in the case of scene MPH16, the model struggles to capture the correct size of the bed due to limited bed instances in the training dataset. Our method performs well on objects like sofas and chairs, outperforming HolisticMesh and yielding comparable results to MOVER.

\subsection{Performance vs. Inference Time:}
In Fig. \ref{fig:inference_time} we present an analysis of performance vs. inference time for all the methods.
Our method (0.75 sec/frame) and Total3D (0.24 sec/frame) \cite{Nie_2020_CVPR} perform per-frame prediction, whereas MOVER \cite{yi2022human} and HolisticMesh \cite{weng2021holistic} perform optimization over a sequence of images. 
Weng \etal \cite{weng2021holistic} take around 4-5 mins/frame and optimize over the whole sequence which takes days to produce the final reconstruction. Whereas, Yi \etal \cite{yi2022human} take around 30 mins to optimize over the whole sequence and take the same time for a scene regardless of the number of frames in the input video. For a fair comparison, the execution time for all the methods is reported over a sequence of frames (avg. 1200 frames). All methods are evaluated under the same system configuration. 
Although the inference time of Total3D \cite{Nie_2020_CVPR} is low, it should be noted that Total3D performs only object reconstruction. Our method achieves comparable performance with MOVER at a much lower per-frame execution time.

\begin{table}[ht]
\centering
    \noindent\resizebox{.8\columnwidth}{!}{
\begin{tabular}{|c|c|c|c|c|}
\hline 
 & \multicolumn{2}{c|}{PROX quantitative} & \multicolumn{2}{c|}{PROX qualitative} \\
\hline  
 Method &  MPJPE & V2V &  MPJPE & V2V\\
\hline
Ours w/o gan & \textbf{181.97} & 198.80 & \textbf{181.74} & 188.17 \\
Ours w/ gan & 181.02 & \textbf{187.89} & 182.57 & \textbf{187.79} \\
\hline
\end{tabular}
}
\caption{\footnotesize{Ablation for human reconstruction module.}}
\label{tab:quantitative_human_abl}
\end{table}
  
\subsection{Ablation Study:}
\noindent
\textbf{Object reconstruction} In Table \ref{tab:quantitative_abl}, we have compared the effect of using the graph discriminator (represented as '$+$ gan' in Table) for the scene reconstruction using 3D and 2D IoU values on the \textit{PROX qualitative} test set. Using the graph discriminator helps in achieving better reconstruction compared to the baseline. Also, we show the effectiveness of using binary object masks as input for object reconstruction. Object masks help in increasing the performance, specifically for the occluded objects, and give the best IoU3D. 

\noindent
\textbf{Human reconstruction} In Table \ref{tab:quantitative_human_abl} we have shown the effectiveness of our scene-aware human body reconstruction on the test set of both \textit{quantitative} and \textit{qualitative} datasets. Using a discriminator to analyze the human placement with respect to the object alignments in the scene helps in achieving better MPJPE and V2V values on \textit{PROX quantitative}.

\section{Conclusions}
We have proposed a fully learning-based scene reconstruction method, which relies on implicit feature representation of a scene for differentiating a physically plausible scene reconstruction from implausible human and object alignments without explicitly defining physical laws and constraints. The reconstruction generator learns from training data to produce a plausible 3D scene and performs per-frame prediction without any test time optimization. Our method achieves comparable performance as the state-of-the-art methods but with a faster reconstruction speed. However, the execution time needs to be further improved for utilizing it in an actual robotics platform. Due to limited variability in the training data, our method suffers from a lack of generalization in terms of dynamic views, camera setup, resolution \etc. It fails to perform well for out-of-training distribution data. In the future, with the availability of a more generalized dataset, robust scene reconstruction would be possible.








%
\bibliographystyle{IEEEtran}
\bibliography{IEEEexample}

\begin{thebibliography}{10}
\providecommand{\url}[1]{#1}
\csname url@rmstyle\endcsname
\providecommand{\newblock}{\relax}
\providecommand{\bibinfo}[2]{#2}
\providecommand\BIBentrySTDinterwordspacing{\spaceskip=0pt\relax}
\providecommand\BIBentryALTinterwordstretchfactor{4}
\providecommand\BIBentryALTinterwordspacing{\spaceskip=\fontdimen2\font plus
\BIBentryALTinterwordstretchfactor\fontdimen3\font minus
  \fontdimen4\font\relax}
\providecommand\BIBforeignlanguage[2]{{%
\expandafter\ifx\csname l@#1\endcsname\relax
\typeout{** WARNING: IEEEtran.bst: No hyphenation pattern has been}%
\typeout{** loaded for the language `#1'. Using the pattern for}%
\typeout{** the default language instead.}%
\else
\language=\csname l@#1\endcsname
\fi
#2}}

\bibitem{zhang2021holistic}
C.~Zhang, Z.~Cui, Y.~Zhang, B.~Zeng, M.~Pollefeys, and S.~Liu, ``Holistic 3d
  scene understanding from a single image with implicit representation,'' in
  \emph{Proceedings of the IEEE/CVF Conference on Computer Vision and Pattern
  Recognition}, 2021, pp. 8833--8842.

\bibitem{choutas2020monocular}
V.~Choutas, G.~Pavlakos, T.~Bolkart, D.~Tzionas, and M.~J. Black, ``Monocular
  expressive body regression through body-driven attention,'' in \emph{European
  Conference on Computer Vision}.\hskip 1em plus 0.5em minus 0.4em\relax
  Springer, 2020, pp. 20--40.

\bibitem{kocabas2020vibe}
M.~Kocabas, N.~Athanasiou, and M.~J. Black, ``Vibe: Video inference for human
  body pose and shape estimation,'' in \emph{Proceedings of the IEEE/CVF
  conference on computer vision and pattern recognition}, 2020, pp. 5253--5263.

\bibitem{kanazawa2018end}
A.~Kanazawa, M.~J. Black, D.~W. Jacobs, and J.~Malik, ``End-to-end recovery of
  human shape and pose,'' in \emph{Proceedings of the IEEE conference on
  computer vision and pattern recognition}, 2018, pp. 7122--7131.

\bibitem{rong2021frankmocap}
Y.~Rong, T.~Shiratori, and H.~Joo, ``Frankmocap: A monocular 3d whole-body pose
  estimation system via regression and integration,'' in \emph{Proceedings of
  the IEEE/CVF International Conference on Computer Vision}, 2021, pp.
  1749--1759.

\bibitem{kocabas2021pare}
M.~Kocabas, C.-H.~P. Huang, O.~Hilliges, and M.~J. Black, ``Pare: Part
  attention regressor for 3d human body estimation,'' in \emph{Proceedings of
  the IEEE/CVF International Conference on Computer Vision}, 2021, pp.
  11\,127--11\,137.

\bibitem{dahnert2021panoptic}
M.~Dahnert, J.~Hou, M.~Nie{\ss}ner, and A.~Dai, ``Panoptic 3d scene
  reconstruction from a single rgb image,'' \emph{Advances in Neural
  Information Processing Systems}, vol.~34, pp. 8282--8293, 2021.

\bibitem{gkioxari2019mesh}
G.~Gkioxari, J.~Malik, and J.~Johnson, ``Mesh r-cnn,'' in \emph{Proceedings of
  the IEEE/CVF International Conference on Computer Vision}, 2019, pp.
  9785--9795.

\bibitem{Nie_2020_CVPR}
Y.~Nie, X.~Han, S.~Guo, Y.~Zheng, J.~Chang, and J.~J. Zhang,
  ``Total3dunderstanding: Joint layout, object pose and mesh reconstruction for
  indoor scenes from a single image,'' in \emph{IEEE/CVF Conference on Computer
  Vision and Pattern Recognition (CVPR)}, June 2020.

\bibitem{song2017semantic}
S.~Song, F.~Yu, A.~Zeng, A.~X. Chang, M.~Savva, and T.~Funkhouser, ``Semantic
  scene completion from a single depth image,'' in \emph{Proceedings of the
  IEEE conference on computer vision and pattern recognition}, 2017, pp.
  1746--1754.

\bibitem{nie2022learning}
Y.~Nie, A.~Dai, X.~Han, and M.~Nie{\ss}ner, ``Learning 3d scene priors with 2d
  supervision,'' \emph{arXiv preprint arXiv:2211.14157}, 2022.

\bibitem{weng2021holistic}
Z.~Weng and S.~Yeung, ``Holistic 3d human and scene mesh estimation from single
  view images,'' in \emph{Proceedings of the IEEE/CVF Conference on Computer
  Vision and Pattern Recognition}, 2021, pp. 334--343.

\bibitem{yi2022human}
H.~Yi, C.-H.~P. Huang, D.~Tzionas, M.~Kocabas, M.~Hassan, S.~Tang, J.~Thies,
  and M.~J. Black, ``Human-aware object placement for visual environment
  reconstruction,'' \emph{arXiv preprint arXiv:2203.03609}, 2022.

\bibitem{kirillov2020pointrend}
A.~Kirillov, Y.~Wu, K.~He, and R.~Girshick, ``Pointrend: Image segmentation as
  rendering,'' in \emph{Proceedings of the IEEE/CVF conference on computer
  vision and pattern recognition}, 2020, pp. 9799--9808.

\bibitem{8765346}
Z.~{Cao}, G.~{Hidalgo Martinez}, T.~{Simon}, S.~{Wei}, and Y.~A. {Sheikh},
  ``Openpose: Realtime multi-person 2d pose estimation using part affinity
  fields,'' \emph{IEEE Transactions on Pattern Analysis and Machine
  Intelligence}, 2019.

\bibitem{corso2020principal}
G.~Corso, L.~Cavalleri, D.~Beaini, P.~Li{\`o}, and P.~Veli{\v{c}}kovi{\'c},
  ``Principal neighbourhood aggregation for graph nets,'' \emph{Advances in
  Neural Information Processing Systems}, vol.~33, pp. 13\,260--13\,271, 2020.

\bibitem{hassan2019resolving}
M.~Hassan, V.~Choutas, D.~Tzionas, and M.~J. Black, ``Resolving 3d human pose
  ambiguities with 3d scene constraints,'' in \emph{Proceedings of the IEEE/CVF
  International Conference on Computer Vision}, 2019, pp. 2282--2292.

\bibitem{rempe2021humor}
D.~Rempe, T.~Birdal, A.~Hertzmann, J.~Yang, S.~Sridhar, and L.~J. Guibas,
  ``Humor: 3d human motion model for robust pose estimation,'' in
  \emph{Proceedings of the IEEE/CVF International Conference on Computer
  Vision}, 2021, pp. 11\,488--11\,499.

\bibitem{shimada2020physcap}
S.~Shimada, V.~Golyanik, W.~Xu, and C.~Theobalt, ``Physcap: Physically
  plausible monocular 3d motion capture in real time,'' \emph{ACM Transactions
  on Graphics (ToG)}, vol.~39, no.~6, pp. 1--16, 2020.

\bibitem{hassan2023synthesizing}
M.~Hassan, Y.~Guo, T.~Wang, M.~Black, S.~Fidler, and X.~B. Peng, ``Synthesizing
  physical character-scene interactions,'' \emph{arXiv preprint
  arXiv:2302.00883}, 2023.

\bibitem{luo2022embodied}
Z.~Luo, S.~Iwase, Y.~Yuan, and K.~Kitani, ``Embodied scene-aware human pose
  estimation,'' \emph{arXiv preprint arXiv:2206.09106}, 2022.

\bibitem{hassan2021populating}
M.~Hassan, P.~Ghosh, J.~Tesch, D.~Tzionas, and M.~J. Black, ``Populating 3d
  scenes by learning human-scene interaction,'' in \emph{Proceedings of the
  IEEE/CVF Conference on Computer Vision and Pattern Recognition}, 2021, pp.
  14\,708--14\,718.

\bibitem{PSI:2019}
\BIBentryALTinterwordspacing
Y.~Zhang, M.~Hassan, H.~Neumann, M.~J. Black, and S.~Tang, ``Generating 3d
  people in scenes without people,'' in \emph{Computer Vision and Pattern
  Recognition (CVPR)}, June 2020. [Online]. Available:
  \url{https://arxiv.org/abs/1912.02923}
\BIBentrySTDinterwordspacing

\bibitem{monszpart2019imapper}
A.~Monszpart, P.~Guerrero, D.~Ceylan, E.~Yumer, and N.~J. Mitra, ``imapper:
  interaction-guided scene mapping from monocular videos,'' \emph{ACM
  Transactions On Graphics (TOG)}, vol.~38, no.~4, pp. 1--15, 2019.

\bibitem{chen2019holistic++}
Y.~Chen, S.~Huang, T.~Yuan, S.~Qi, Y.~Zhu, and S.-C. Zhu, ``Holistic++ scene
  understanding: Single-view 3d holistic scene parsing and human pose
  estimation with human-object interaction and physical commonsense,'' in
  \emph{Proceedings of the IEEE/CVF International Conference on Computer
  Vision}, 2019, pp. 8648--8657.

\bibitem{zhang2020perceiving}
J.~Y. Zhang, S.~Pepose, H.~Joo, D.~Ramanan, J.~Malik, and A.~Kanazawa,
  ``Perceiving 3d human-object spatial arrangements from a single image in the
  wild,'' in \emph{European Conference on Computer Vision}.\hskip 1em plus
  0.5em minus 0.4em\relax Springer, 2020, pp. 34--51.

\bibitem{xie2022chore}
X.~Xie, B.~L. Bhatnagar, and G.~Pons-Moll, ``Chore: Contact, human and object
  reconstruction from a single rgb image,'' \emph{CVPR}, 2022.

\bibitem{dabral2021gravity}
R.~Dabral, S.~Shimada, A.~Jain, C.~Theobalt, and V.~Golyanik, ``Gravity-aware
  monocular 3d human-object reconstruction,'' \emph{arXiv preprint
  arXiv:2108.08844}, 2021.

\bibitem{li2022cliff}
Z.~Li, J.~Liu, Z.~Zhang, S.~Xu, and Y.~Yan, ``Cliff: Carrying location
  information in full frames into human pose and shape estimation,''
  \emph{arXiv preprint arXiv:2208.00571}, 2022.

\bibitem{loper2015smpl}
M.~Loper, N.~Mahmood, J.~Romero, G.~Pons-Moll, and M.~J. Black, ``Smpl: A
  skinned multi-person linear model,'' \emph{ACM transactions on graphics
  (TOG)}, vol.~34, no.~6, pp. 1--16, 2015.

\bibitem{hu2018relation}
H.~Hu, J.~Gu, Z.~Zhang, J.~Dai, and Y.~Wei, ``Relation networks for object
  detection,'' in \emph{Proceedings of the IEEE conference on computer vision
  and pattern recognition}, 2018, pp. 3588--3597.

\bibitem{kolotouros2019learning}
N.~Kolotouros, G.~Pavlakos, M.~J. Black, and K.~Daniilidis, ``Learning to
  reconstruct 3d human pose and shape via model-fitting in the loop,'' in
  \emph{Proceedings of the IEEE/CVF International Conference on Computer
  Vision}, 2019, pp. 2252--2261.

\bibitem{he2016deep}
K.~He, X.~Zhang, S.~Ren, and J.~Sun, ``Deep residual learning for image
  recognition,'' in \emph{Proceedings of the IEEE conference on computer vision
  and pattern recognition}, 2016, pp. 770--778.

\bibitem{pavlakos2019expressive}
G.~Pavlakos, V.~Choutas, N.~Ghorbani, T.~Bolkart, A.~A. Osman, D.~Tzionas, and
  M.~J. Black, ``Expressive body capture: 3d hands, face, and body from a
  single image,'' in \emph{Proceedings of the IEEE/CVF conference on computer
  vision and pattern recognition}, 2019, pp. 10\,975--10\,985.

\bibitem{mahmood2019amass}
N.~Mahmood, N.~Ghorbani, N.~F. Troje, G.~Pons-Moll, and M.~J. Black, ``Amass:
  Archive of motion capture as surface shapes,'' in \emph{Proceedings of the
  IEEE/CVF international conference on computer vision}, 2019, pp. 5442--5451.

\bibitem{pcaref}
K.~P. F.R.S., ``Liii. on lines and planes of closest fit to systems of points
  in space,'' \emph{The London, Edinburgh, and Dublin Philosophical Magazine
  and Journal of Science}, vol.~2, no.~11, pp. 559--572, 1901.

\bibitem{FeyLenssen2019}
M.~Fey and J.~E. Lenssen, ``Fast graph representation learning with {PyTorch
  Geometric},'' in \emph{ICLR Workshop on Representation Learning on Graphs and
  Manifolds}, 2019.

\bibitem{kingma2014adam}
D.~P. Kingma and J.~Ba, ``Adam: A method for stochastic optimization,''
  \emph{arXiv preprint arXiv:1412.6980}, 2014.

\end{thebibliography}

\end{document}